\documentclass[sigconf]{acmart}

\usepackage{booktabs} 
\usepackage{graphicx,balance} 
\usepackage{caption,subcaption}
\usepackage{indentfirst}

\setcopyright{rightsretained}

\acmDOI{10.475/123_4}

\acmISBN{123-4567-24-567/08/06}

\acmYear{2017}
\copyrightyear{2017}

\acmArticle{X}
\acmPrice{15.00}

\begin{document}
\title{Discovering Hidden Structure in High Dimensional Human \\ Behavioral Data via Tensor Factorization}

\author[H. Hosseinmardi]{Homa Hosseinmardi}
\affiliation{%
  \institution{USC Information Sciences Institute}
  \city{Marina del Rey}
  \state{CA}
  \postcode{90292}
}
\email{homahoss@isi.edu}

\author[H. Kao]{Hsien-Te Kao}
\affiliation{%
  \institution{USC Information Sciences Institute}
  \city{Marina del Rey}
  \state{CA}
  \postcode{90292}
}
\email{hsientek@usc.edu}

\author[K. Lerman]{Kristina Lerman}
\affiliation{%
  \institution{USC Information Sciences Institute}
  \city{Marina del Rey}
  \state{CA}
  \postcode{90292}
}
\email{klerman@isi.edu}

\author[E. Ferrara]{Emilio Ferrara}
\affiliation{%
  \institution{USC Information Sciences Institute}
  \city{Marina del Rey}
  \state{CA}
  \postcode{90292}
}
\email{emiliofe@usc.edu}

\begin{abstract}
In recent years, the rapid growth in technology has increased the opportunity for longitudinal human behavioral studies. Rich multimodal data, from wearables like Fitbit, online social networks, mobile phones etc. can be collected in natural environments. Uncovering the underlying low-dimensional structure of noisy multi-way data in an unsupervised setting is a challenging problem. Tensor factorization has been  successful in extracting the interconnected low-dimensional descriptions of multi-way data. In this paper, we apply non-negative tensor factorization on a real-word wearable sensor data, \textit{StudentLife}, to find latent temporal factors and group of similar individuals. Meta data is available for the semester schedule, as well as  the individuals' performance and personality. We demonstrate that non-negative tensor factorization can successfully discover clusters of individuals who exhibit higher academic performance, as well as those who frequently engage in leisure activities. The recovered latent temporal patterns associated with these groups are validated against ground truth data to demonstrate the accuracy of our framework.

\end{abstract}

%
%


\maketitle

\section{Introduction}
	Behavioral data, collected from a variety of sources, has been used to understand human affect, wellbeing, social relationships, performance and decision making progress. Existing techniques such as interviews and questionnaires are inaccurate, expensive and laborious to administer. Fortunately, today's densely instrumented world offers tremendous opportunities for continuous acquisition and analysis of multi-variant time series data that provides a multimodal, spatiotemporal characterization of an individual's actions. 
    
 Efficiently coupling such rich sensor data with fusion and predictive modeling techniques can provide continuous, personalized, contextual, and insightful assessments of individual performance. StudentLife is a 10-week study on 48 Dartmouth undergraduate and graduate students using passive and mobile sensor data to infer wellbeing, academic performance and behavioral trends \cite{3-1}. The Dartmouth College research team was able to predict GPA using activity, conversational interaction, mobility, and self-reported emotion and stress data over the semester. SNAPSHOT is a 30-day study on MIT undergraduates using mobile sensors and surveys to understand sleep, social interactions, affect, performance, stress and health \cite{3-2}. 
 RealityMining is a 9-month study on 75 MIT Media Laboratory students, 
 using mobile sensor data to track the social interactions and networkings \cite{3-3}. The friends-and-families study collects data from 130 adult members of a young family community to study fitness intervention and social incentives \cite{aharony2011social}. 
 
 These data sources are often collected from small set of participants in natural condition, continuously and over long periods of time. Therefore, they are potentially heterogeneous, sparse, in high dimensional regime and have systematically missing values. Tensor factorization can tolerate missing values~\cite{4-1,4-2,4-3} and has been used popularly in multi-way relational data, for example clustering and temporal structure discovery in dynamic networks~\cite{sapienza2017non}. In this paper we demonstrate non-negative tensor factorization (NTF) can reveal the low-dimensional patterns of noisy heterogeneous wearable sensor data, while preserving the interpretability \cite{cichocki2009nonnegative}. We are mainly interested in StudentLife dataset as it has a very rich set of time series collected from a cohort of students via their smartphones, plus wide range of meta data available, such as workloads and mental health state. Ref. \cite{wang2015smartgpa} builds a matrix representation of this longitudinal data by extracting different statistics from variables across the time dimension to infer students' wellbeing and performance. Pattern of activity and sociability behavior has been explored in \cite{harari2017patterns}, by aggregating variables over the individuals dimension. We would like to extract the interconnected low-dimensional latent factors, without aggregation of data over any dimension. Here we will demonstrate the strength of tensor factorization for \textit{in situ} human behavior research through finding the hidden structures and validate them against ground truth. First we evaluate the variables and individuals associated with each temporal latent factor. Then we compare the distribution of students who are strongly associated with each component, across different meta data, such as GPA, personality traits and affect.

\subsection*{Contributions}
In this article we propose the following ideas:

\begin{itemize}
  \item Unsupervised extraction of low-dimensional structure of wearable sensor data
  \item Discovering clusters of individuals who exhibit higher academic performance, as well as those who frequently engage in leisure activities
  
  \item Validation of recovered latent temporal patterns associated with these groups against ground truth data

\end{itemize}

\section{Methodology}
\subsection{Preliminaries}
A wide range of real-world datasets such as recommendation system, multivariate time series and video streams  are multi-way. A mathematical representation of such data is tensor, $\mathcal{X} \in \mathbb{R}^{I_1 \times ... \times I_N}$. One approach to work with this multi-way data is matrixizing or flattening and applying conventional supervised/unsupervised techniques, but we may loose structural information. Tensors are powerful tools to extract complex structures from the high-dimensional multi-way data in an unsupervised approach. In this section we discuss the model that we will use to analyze a 3-way longitudinal human behavioral data. For $I$ individuals, with $J$ variables for a duration of $K$ time frames, tensor $\mathcal{X} \in \mathbb{R}^{I \times J \times K}$ will be created. Entry $x_{ijk}$ of this tensor corresponds to the $i^{th}$ person in the $j^{th}$ variable at the $k^{th}$ time unit. Table \ref{tab:notation} summarizes our notation throughout this paper. 
\begin{figure*}[h]
    \centering
    \includegraphics[width=0.65\textwidth]{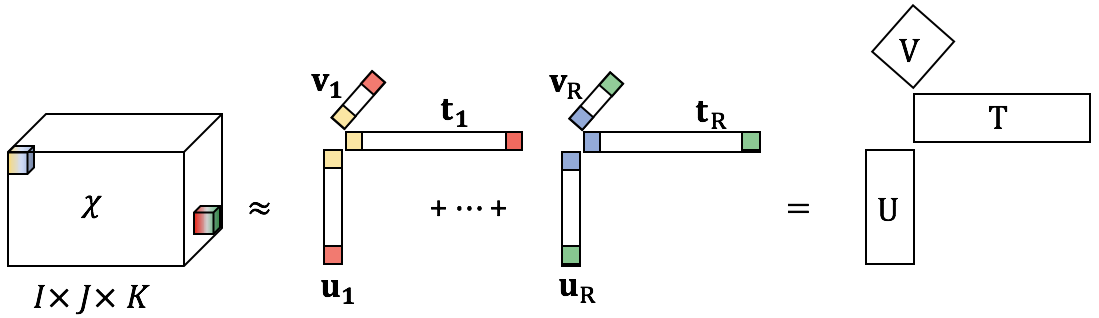}
    \caption{The CANDECOMP/PARAFAC decomposition.}
    \label{fig:cp}
\end{figure*}

\begin{table}
	\centering
    \caption{Table of Symbols.}
	\begin{tabular}{c|c}
	\hline \hline
	Symbol & Definition \\ \hline
	\hline
    $\mathcal{X},\mathrm{X},\mathbf{x},\mathrm{x}$ & Tensor, matrix, column vector, scaler\\ \hline
	$ \mathbf{x} \in \mathbb{R}^{I\times 1} $ & Definition of an I-dimensional vector  \\	\hline
     $\circ$  & Outer product \\\hline
    $||.||$ & Frobenius norm \\ \hline \hline
	\end{tabular}
	
	\label{tab:notation}  
\end{table}

\subsection{Non-negative tensor decomposition}
  The extraction of meaningful patterns of behavior can be carried out by taking full advantage of tensor decomposition techniques. Let us consider a dataset composed by individuals, whose daily behavior (walking, running, talking, etc. during the day)  has been recorded over time. To uncover groups of individuals with similar correlated trajectories, identification of lower-dimensional factors is required. Once the dataset is represented in the tensor form, we can perform tensor decomposition to discover the hidden lower dimension structure of the data.

Here, we use the CANDECOMP/PARAFAC decomposition \cite{carroll1970analysis,harshman1970foundations}, which will decompose the tensor $\mathcal{X} \in R^{I \times J \times K}$ into sum of rank-one tensors, called components, Figure \ref{fig:cp}. Also we add the non-negativity constraints to each factor matrix in order to find interpretable components.


\begin{align*}
\mathcal{X} \approx \hat{\mathcal{X} } =  \sum_{r=1}^{R} \lambda_r \mathbf{u}_r \circ \mathbf{v}_r \circ \mathbf{t}_r 
\end{align*}

where $\lambda_r$ are the values of the tensor core $L = diag(\Lambda)$, and the outer product $\mathbf{u}_r~o~\mathbf{v}_r~o~\mathbf{t}_r$ corresponds to the $r^{th}$ component of rank-R estimation.  This decomposition can be written as the following optimization problem:
\begin{align*}
\min_{\mathbf{\lambda}, \mathrm{U}, \mathrm{V}, \mathrm{T}} || \mathcal{X} - \hat{\mathcal{X} }||
\\
\mathrm{s.t.}, \mathbf{\lambda}, \mathrm{U}, \mathrm{V}, \mathrm{T} \geq 0
\end{align*}

$\mathbf{U}, \mathrm{V}, \mathrm{T}$ are the factor matrices with their columns containing rank-1 factors, $\mathbf{u}_r, \mathbf{v}_r$, and $\mathbf{t}_r$, respectively. Imposing the non-negativity constraints makes the factorization results interpretable. 


\section{Case Study}
	In this paper we use StudentLife dataset, a large publicly available dataset, tracking student performance, wellbeing and physiological state \cite{2-1}. StudentLife is a 10-week study conducted during 2013 spring semester on 48 Dartmouth students (30 undergraduate and 18 graduate students). The dataset can be divided into four sections: smartphone sensors, ecological momentary assessments (EMAs), psychometrics and academic performances. From raw sensor data, activity (stationary, walk, run and unknown), audio (silence, voice, noise and unknown), and conversation has been inferred. 
    Psychometrics have pre-post Big Five personality \cite{2-2}, flourishing scale \cite{2-3}, UCLA loneliness scale \cite{2-4}, positive and negative affect schedule (PANAS) \cite{2-8}, perceived stress scale (PSS) \cite{2-7}, PHQ-9 depression scale \cite{2-6}, Pittsburgh sleep quality index (PSQI) \cite{2-5} and VR-12 heath scale \cite{2-9}. Academic performances have class schedule, number of deadlines, overall GPA, online class forum Piazza participation, and more. 
    We will not use EMAs in our modeling as we are interested in learning from passively recorded data and we keep survey and EMAs only as ground truth for validation task. For psychometrics surveys, there are large amount of missing pre and post survey scores. We will use the post survey score if it is available otherwise pre survey score is used as replacement. If both scores are missing for a survey, we will drop that users when using that survey. 

\subsection{Feature Extraction}

Our goal in this paper is understanding whether we can find the low-dimensional structure of daily life from wearable devices, without using any meta data such as self-reported EMAs or day of the week. Therefore, we only use the smartphone sensor variables in our feature set. Each time unit comprise one day worth of data, and is divided into four time bins, bedtime (midnight-6 am), morning (6 am-12 pm), afternoon (12 pm-6 pm), and evening (6 pm-midnight). We extract duration (minutes) of running, walking, stationary, silence, voice, noise, and dark, per time-bin in each day. Frequency of each behavior and number of changes in each behavior (e.g. from walking to running) for each time-bin has been also captured. From GPS and WiFi, number of unique locations visited, and from Bluetooth number of unique near by devices per time-bin are added to the variable set. We normalize all the variables to have the same range $[0,1]$ to avoid variables with large values (e.g. duration in minutes) dominate the analysis. At the end, we organize our data as tensor $\mathcal{X}$ with $I=48$ individuals, $J=85$ variables and $K=66$ days. Only 5\% of the tensor is missing, which we imputed by filling them with the mean value.

\section{Discovering pattern of temporal behavior}

\begin{figure}
    \centering
    \includegraphics[width=0.45\textwidth]{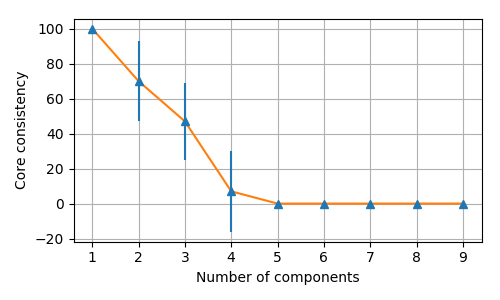}
    \caption{Core consistency curve.}
    \label{fig:cc}
    \vspace{-.5cm}
\end{figure}

In this section we employ non-negative tensor decomposition to find the hidden structure in the data. However, we need to find and fixate the correct number of components $R$. To ensure of selecting the best approximation, we change the number of components $R=1,2,..,9$ and report the mean and standard deviation of the core consistency scores for each $R$, Figure \ref{fig:cc}. Negative values are imputed to zero and no standard deviation is reported for them. We choose rank $R=3$, the number of components that yields the largest change in the slope of the core consistency curve. After fixing the number of components, we then select the best approximation after a set of random initializations, by choosing the one corresponding to the maximum value of core consistency for the selected rank $R=3$. 

\begin{figure}
    \centering
    \includegraphics[width=0.25\textwidth]{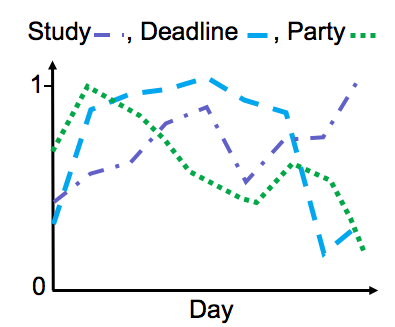}
    \caption{Students deadline, party and study trends over the semester, obtained from StudenLife website~\cite{studentlife}. All curves are normalized to one.}
    \label{fig:gt}
\end{figure}

\begin{figure*}[htb]
\centering 
\begin{subfigure}{0.33\textwidth}
  \includegraphics[width=\linewidth]{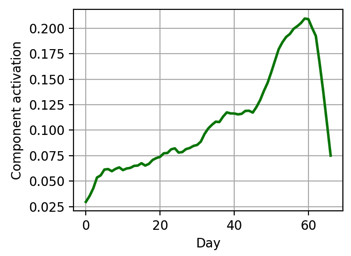}
  \caption{$1^{st}$ component: Studying pattern.}
  \label{fig:uc1}
\end{subfigure}
\begin{subfigure}{0.33\textwidth}
  \includegraphics[width=\linewidth]{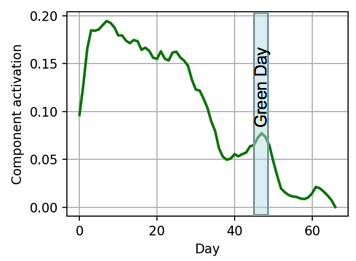}
  \caption{$2^{nd}$ component: Partying pattern.}
  \label{fig:uc2}
\end{subfigure}\hfil 
\begin{subfigure}{0.33\textwidth}
  \includegraphics[width=\linewidth]{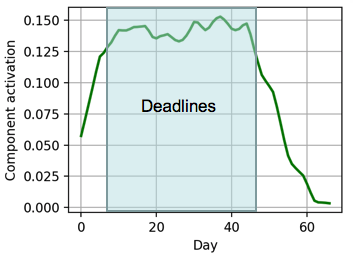}
  \caption{$3^{rd}$ component: Deadline pattern.}
  \label{fig:uc3}
\end{subfigure}\hfil 
\caption{Discovered temporal structure over the semester.  }
\label{fig:all_tcs}
\end{figure*}

Figure \ref{fig:all_tcs} presents the extracted latent temporal components. We will use the meta data available in StudentLife dataset to find the topic of each latent component discovered, Figure~\ref{fig:gt} \cite{studentlife}. This curves are generated based on self-reported values by students and averaged across all students each day. The first component follows the same pattern as studying and increases over the semester, Figure \ref{fig:uc1}. The second component should be related to partying, as it decreases after the first week of semester and there is a jump around the Green Key weekend (Figure \ref{fig:uc2}, light green box). The third component follows the pattern of deadlines. The light green box in Figure \ref{fig:uc3} shows the duration with higher number of deadlines from meta data, Figure\ref{fig:gt}.

\begin{figure*}
    \centering
    \includegraphics[width=0.8\textwidth]{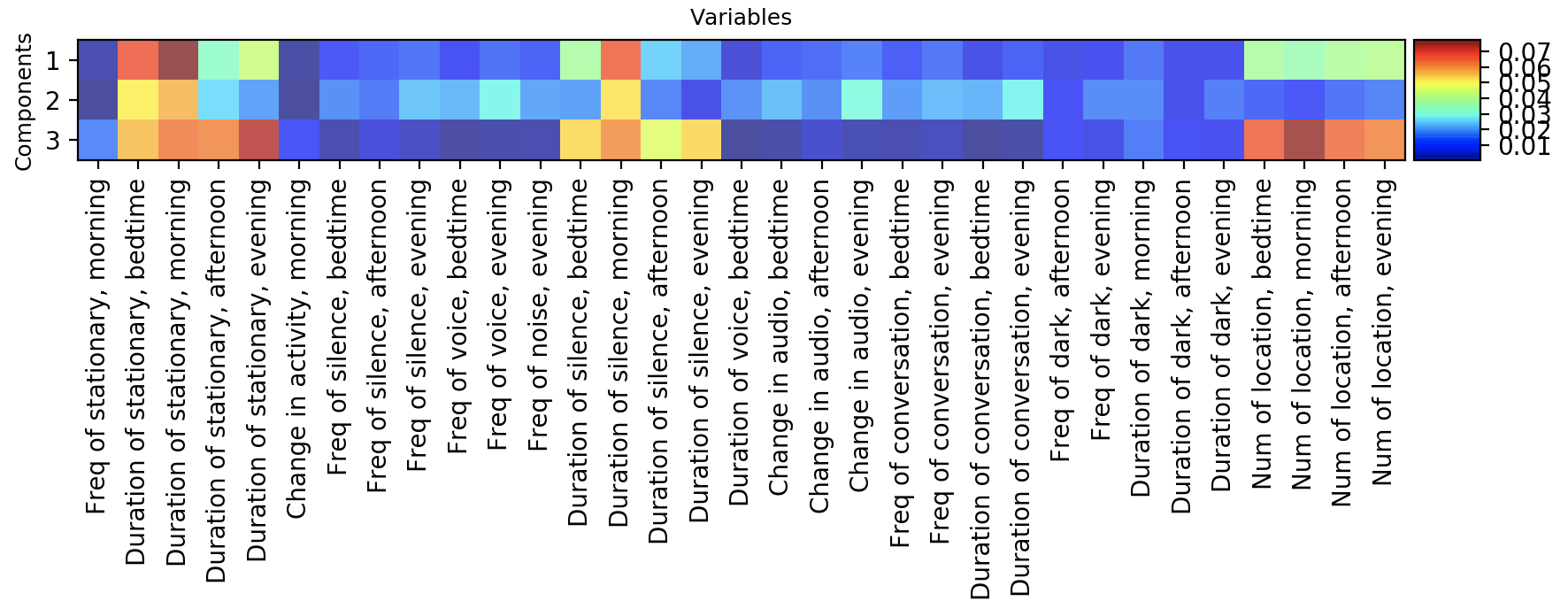}
    \caption{Membership of variables in each of the three components; we sort the variables based on their weights and create the top-weighted variable set in each component. Variables presented here are among the top-weighted ones at least in one of the components. }
    \label{fig:mm}
\end{figure*}

\subsection{Structural Validation}
\begin{figure*}[htb]
    \centering 
\begin{subfigure}{0.3\textwidth}
  \includegraphics[width=\linewidth]{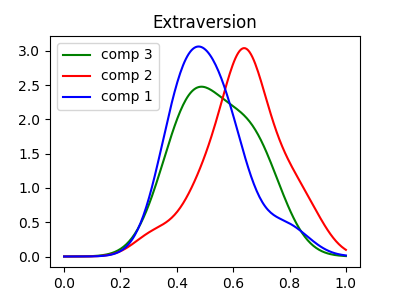}
  \caption{}
  \label{fig:ext}
\end{subfigure}\hfil 
\begin{subfigure}{0.3\textwidth}
  \includegraphics[width=\linewidth]{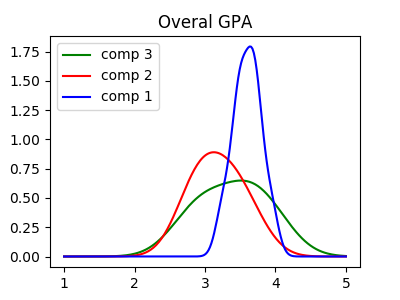}
  \caption{ }
  \label{fig:gpa}
\end{subfigure}\hfil 
\begin{subfigure}{0.3\textwidth}
  \includegraphics[width=\linewidth]{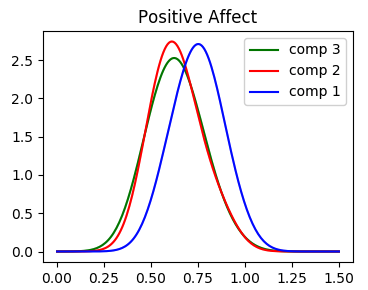}
  \caption{ }
  \label{fig:pa}
\end{subfigure}

\caption{Kernel Density Estimation (KDE) computed on the values related to each personality for top 25\% individuals in each component.  }
\label{fig:all_kde}
\end{figure*}

A nice aspect of the dataset we use is the existence of ground truth available from known semester schedules, students' course load and self-reported values. Here, we examine our hypothesis that extracted components represent studying, partying, and deadline topics. Figure \ref{fig:mm} shows stronger correlation of component 3 with stationary, silence and number of locations visited, which follows the pattern of working on homework deadlines for students. More silence and darkness in the morning, and stronger membership of conversation during afternoon and evening, supports the hypothesis of considering topic "Party" for component 2. Frequency of capturing voice in the evening and number of changes of audio status in afternoon have higher weights in component 2 compare to the other two components. Also we looked at the distribution of top 25\% individuals who have the highest association with each component across different existing meta data, such as personality and performance. Note that all students have some degree of association in all components and one student can fall among top 25\% individuals in more than one component. Kernel Density Estimation (KDE) for the distribution of extraversion scores for students associated to each component is demonstrated in Figure \ref{fig:ext}. Top members of component 2 have significantly bigger mean value than component 1 (p-value=0.01 based on t-test). People who score high in extraversion personality trait, enjoy being with others, participating in social gatherings and partying. This observation is along with our hypothesis which component 2 reflect partying, Figure\ref{fig:uc2}.

Running the ANOVA test for all the other four personality traits (Openness, Conscientiousness, Agreeableness, Neuroticism), the null hypothesis of equal mean values for the three sample sets is not rejected. Component 3 mostly matches the deadline pattern in Figure \ref{fig:gt}. We looked at the average number of deadlines over 10 weeks for the top members in each component and we observed students with higher weight in component 3 have more deadlines than top members of components 1 (p-value = 0.06). 

Looking at the students' performance, it is also very interesting that the individuals involved in component 2 have lower GPA than the other two components, (p-value = 0.008), Figure \ref{fig:gpa}. Based on the average self-reported relaxing by students over the semester, top member of component 1 have significantly less relaxing duration and at the same time, they have the highest contribution in online class forum Piazza participation. These students have significantly more positive affect score compare to group of students associated in the other two components (p-value = 0.03), Figure \ref{fig:pa}.

\section{Conclusions and Future work}

Rich multimodal data collected from wearables devices (e.g. Fitbit), mobile phones, online social networks, etc. become increasingly available to reconstruct digital trails and study human behavior.
In this paper, we adopt the \textit{StudentLife} dataset collected over the course of 10 weeks from 48 Dartmouth undergraduate and graduate students using passive and mobile sensors, with the goal of inferring wellbeing, academic performance, and behavioral trends.
We employ the unsupervised learning framework based on \textit{non-negative tensor decomposition} to find groups of individuals with similar behavior.
This type of decomposition can uncover latent temporal structures such as studying and partying over the semester. 
By applying this framework we could discover traits like  that the group of students associated with \textit{Component 2} (leisure activities) have the highest average scores of self-reported extroversion, while students associated with \textit{Component 1} (studying factor) have the highest average GPA, lower relax-time and higher positive affect. 
Next, we plan to implement supervised predictions of individuals' performance and personality directly from tensors, instead of mapping the tensors to matrices and using conventional supervised methods designed for two dimensional data.

\newpage
\section*{Acknowledgements}

The research is based upon work supported by the Office of the Director of National Intelligence (ODNI), Intelligence Advanced Research Projects Activity (IARPA), via IARPA Contract No 2017-17042800005, and by DARPA (grant no. D16AP00115). The views and conclusions contained herein are those of the authors and should not be interpreted as necessarily representing the official policies or endorsements, either expressed or implied, of the ODNI, IARPA, or the U.S. Government. The U.S. Government is authorized to reproduce and distribute reprints for Governmental purposes notwithstanding any copyright annotation thereon.

\bibliographystyle{ACM-Reference-Format}
\balance
\bibliography{bibliography} 


\begin{thebibliography}{23}


\ifx \showCODEN    \undefined \def \showCODEN     #1{\unskip}     \fi
\ifx \showDOI      \undefined \def \showDOI       #1{#1}\fi
\ifx \showISBNx    \undefined \def \showISBNx     #1{\unskip}     \fi
\ifx \showISBNxiii \undefined \def \showISBNxiii  #1{\unskip}     \fi
\ifx \showISSN     \undefined \def \showISSN      #1{\unskip}     \fi
\ifx \showLCCN     \undefined \def \showLCCN      #1{\unskip}     \fi
\ifx \shownote     \undefined \def \shownote      #1{#1}          \fi
\ifx \showarticletitle \undefined \def \showarticletitle #1{#1}   \fi
\ifx \showURL      \undefined \def \showURL       {\relax}        \fi
\providecommand\bibfield[2]{#2}
\providecommand\bibinfo[2]{#2}
\providecommand\natexlab[1]{#1}
\providecommand\showeprint[2][]{arXiv:#2}

\bibitem[\protect\citeauthoryear{Acar, Dunlavy, Kolda, and Morup}{Acar
  et~al\mbox{.}}{2011}]%
        {4-1}
\bibfield{author}{\bibinfo{person}{Evrim Acar}, \bibinfo{person}{Daniel~M.
  Dunlavy}, \bibinfo{person}{Tamara~G. Kolda}, {and} \bibinfo{person}{Morten
  Morup}.} \bibinfo{year}{2011}\natexlab{}.
\newblock \showarticletitle{Scalable Tensor Factorizations for Incomplete
  Data}.
\newblock \bibinfo{journal}{{\em Chemometrics and Intelligent Laboratory
  Systems\/}} \bibinfo{volume}{106}, \bibinfo{number}{1} (\bibinfo{date}{March}
  \bibinfo{year}{2011}), \bibinfo{pages}{41--56}.
\newblock
\showDOI{%
\url{https://doi.org/10.1016/j.chemolab.2010.08.004}}


\bibitem[\protect\citeauthoryear{Aharony, Pan, Ip, Khayal, and
  Pentland}{Aharony et~al\mbox{.}}{2011}]%
        {aharony2011social}
\bibfield{author}{\bibinfo{person}{Nadav Aharony}, \bibinfo{person}{Wei Pan},
  \bibinfo{person}{Cory Ip}, \bibinfo{person}{Inas Khayal}, {and}
  \bibinfo{person}{Alex Pentland}.} \bibinfo{year}{2011}\natexlab{}.
\newblock \showarticletitle{Social fMRI: Investigating and shaping social
  mechanisms in the real world}.
\newblock \bibinfo{journal}{{\em Pervasive and Mobile Computing\/}}
  \bibinfo{volume}{7}, \bibinfo{number}{6} (\bibinfo{year}{2011}),
  \bibinfo{pages}{643--659}.
\newblock


\bibitem[\protect\citeauthoryear{Buysse, Reynolds, Monk, Berman, and
  Kupfer}{Buysse et~al\mbox{.}}{1989}]%
        {2-5}
\bibfield{author}{\bibinfo{person}{Daniel~J. Buysse}, \bibinfo{person}{C.F.
  Reynolds}, \bibinfo{person}{T.H. Monk}, \bibinfo{person}{S.R. Berman}, {and}
  \bibinfo{person}{D.J. Kupfer}.} \bibinfo{year}{1989}\natexlab{}.
\newblock \showarticletitle{The Pittsburgh Sleep Quality Index (PSQI): A New
  Instrument for Psychiatric Research and Practice}.
\newblock \bibinfo{journal}{{\em Psychiatry Research\/}} \bibinfo{volume}{28},
  \bibinfo{number}{2} (\bibinfo{year}{1989}), \bibinfo{pages}{193--213}.
\newblock
\showURL{%
\url{http://uacc.arizona.edu/sites/default/files/psqi_sleep_questionnaire_1_pg.pdf}}


\bibitem[\protect\citeauthoryear{Carroll and Chang}{Carroll and Chang}{1970}]%
        {carroll1970analysis}
\bibfield{author}{\bibinfo{person}{J~Douglas Carroll} {and}
  \bibinfo{person}{Jih-Jie Chang}.} \bibinfo{year}{1970}\natexlab{}.
\newblock \showarticletitle{Analysis of individual differences in
  multidimensional scaling via an N-way generalization of “Eckart-Young”
  decomposition}.
\newblock \bibinfo{journal}{{\em Psychometrika\/}} \bibinfo{volume}{35},
  \bibinfo{number}{3} (\bibinfo{year}{1970}), \bibinfo{pages}{283--319}.
\newblock


\bibitem[\protect\citeauthoryear{Cichocki}{Cichocki}{2009}]%
        {cichocki2009nonnegative}
\bibfield{author}{\bibinfo{person}{Andrzej Cichocki}.}
  \bibinfo{year}{2009}\natexlab{}.
\newblock \bibinfo{booktitle}{{\em Nonnegative matrix and tensor
  factorizations: applications to exploratory multi way data analysis and blind
  source separation}}.
\newblock \bibinfo{publisher}{John Wiley and Sons}.
\newblock


\bibitem[\protect\citeauthoryear{Cichocki}{Cichocki}{2014}]%
        {4-3}
\bibfield{author}{\bibinfo{person}{Andrzej Cichocki}.}
  \bibinfo{year}{2014}\natexlab{}.
\newblock \showarticletitle{Era of Big Data Processing: A New Approach via
  Tensor Networks and Tensor Decompositions}.
\newblock \bibinfo{journal}{{\em Computing Research Repository\/}}
  (\bibinfo{date}{August} \bibinfo{year}{2014}), \bibinfo{pages}{1--30}.
\newblock
\showURL{%
\url{https://arxiv.org/abs/1403.2048}}


\bibitem[\protect\citeauthoryear{Cohen, Kamarck, and Mermelstein}{Cohen
  et~al\mbox{.}}{1983}]%
        {2-7}
\bibfield{author}{\bibinfo{person}{S. Cohen}, \bibinfo{person}{T. Kamarck},
  {and} \bibinfo{person}{R. Mermelstein}.} \bibinfo{year}{1983}\natexlab{}.
\newblock \showarticletitle{A Global Measure of Perceived Stress}.
\newblock \bibinfo{journal}{{\em Journal of Health and Social Behaviour\/}}
  \bibinfo{volume}{24}, \bibinfo{number}{4} (\bibinfo{year}{1983}),
  \bibinfo{pages}{385--396}.
\newblock
\showURL{%
\url{https://das.nh.gov/wellness/Docs/Percieved%20Stress%20Scale.pdf}}


\bibitem[\protect\citeauthoryear{Dauwels, Garg, Earnest, and Pang}{Dauwels
  et~al\mbox{.}}{2012}]%
        {4-2}
\bibfield{author}{\bibinfo{person}{Justin Dauwels}, \bibinfo{person}{Lalit
  Garg}, \bibinfo{person}{Arul Earnest}, {and} \bibinfo{person}{Leong~K.
  Pang}.} \bibinfo{year}{2012}\natexlab{}.
\newblock \showarticletitle{Tensor Factorization for Missing Data Imputation in
  Medical Questionnaires}.
\newblock \bibinfo{journal}{{\em 2012 IEEE International Conference on
  Acoustics, Speech and Signal Processing\/}} (\bibinfo{date}{March}
  \bibinfo{year}{2012}), \bibinfo{pages}{2109--2112}.
\newblock
\showDOI{%
\url{https://doi.org/10.1109/ICASSP.2012.6288327}}


\bibitem[\protect\citeauthoryear{Diener, Wirtz, Tov, Kim-Prieto, Choi, Oishi,
  and Biswas-Diener}{Diener et~al\mbox{.}}{2009}]%
        {2-3}
\bibfield{author}{\bibinfo{person}{Ed Diener}, \bibinfo{person}{Derrick Wirtz},
  \bibinfo{person}{William Tov}, \bibinfo{person}{Chu Kim-Prieto},
  \bibinfo{person}{Dong-won Choi}, \bibinfo{person}{Shigehiro Oishi}, {and}
  \bibinfo{person}{Robert Biswas-Diener}.} \bibinfo{year}{2009}\natexlab{}.
\newblock \showarticletitle{New Well-being Measures: Short Scales to Assess
  Flourishing and Positive and Negative Feelings}.
\newblock \bibinfo{journal}{{\em Social Indicators Research\/}}
  \bibinfo{volume}{39} (\bibinfo{year}{2009}), \bibinfo{pages}{247--266}.
\newblock
\showURL{%
\url{https://internal.psychology.illinois.edu/~ediener/Documents/FS.pdf}}


\bibitem[\protect\citeauthoryear{Dror, Andrew, Fanglin, Zhenyu, Tianxing, Rui,
  Xia, Gabriella, and Stefanie}{Dror et~al\mbox{.}}{2014}]%
        {studentlife}
\bibfield{author}{\bibinfo{person}{Ben-Zeev Dror}, \bibinfo{person}{Campbell
  Andrew}, \bibinfo{person}{Chen Fanglin}, \bibinfo{person}{Chen Zhenyu},
  \bibinfo{person}{Li Tianxing}, \bibinfo{person}{Wang Rui},
  \bibinfo{person}{Zhou Xia}, \bibinfo{person}{Harari Gabriella}, {and}
  \bibinfo{person}{Tignor Stefanie}.} \bibinfo{year}{2014}\natexlab{}.
\newblock \bibinfo{title}{StudentLife Dataset - Dartmouth College}.
\newblock \bibinfo{howpublished}{\url{http://studentlife.cs.dartmouth.edu}}.
  (\bibinfo{year}{2014}).
\newblock


\bibitem[\protect\citeauthoryear{Eagle and Pentland}{Eagle and
  Pentland}{2006}]%
        {3-3}
\bibfield{author}{\bibinfo{person}{Nathan Eagle} {and} \bibinfo{person}{Alex
  Pentland}.} \bibinfo{year}{2006}\natexlab{}.
\newblock \showarticletitle{Reality Mining: Sensing Complex Social Systems}.
\newblock \bibinfo{journal}{{\em Personal and Ubiquitous Computing\/}}
  \bibinfo{volume}{10}, \bibinfo{number}{4} (\bibinfo{date}{May}
  \bibinfo{year}{2006}), \bibinfo{pages}{255--268}.
\newblock
\showDOI{%
\url{https://doi.org/10.1007/s00779-005-0046-3}}


\bibitem[\protect\citeauthoryear{Harari, Gosling, Wang, Chen, Chen, and
  Campbell}{Harari et~al\mbox{.}}{2017}]%
        {harari2017patterns}
\bibfield{author}{\bibinfo{person}{Gabriella~M Harari},
  \bibinfo{person}{Samuel~D Gosling}, \bibinfo{person}{Rui Wang},
  \bibinfo{person}{Fanglin Chen}, \bibinfo{person}{Zhenyu Chen}, {and}
  \bibinfo{person}{Andrew~T Campbell}.} \bibinfo{year}{2017}\natexlab{}.
\newblock \showarticletitle{Patterns of behavior change in students over an
  academic term: A preliminary study of activity and sociability behaviors
  using smartphone sensing methods}.
\newblock \bibinfo{journal}{{\em Computers in Human Behavior\/}}
  \bibinfo{volume}{67} (\bibinfo{year}{2017}), \bibinfo{pages}{129--138}.
\newblock


\bibitem[\protect\citeauthoryear{Harshman}{Harshman}{1970}]%
        {harshman1970foundations}
\bibfield{author}{\bibinfo{person}{Richard~A Harshman}.}
  \bibinfo{year}{1970}\natexlab{}.
\newblock \showarticletitle{Foundations of the parafac procedure: models and
  conditions for an" explanatory" multimodal factor analysis}.
\newblock  (\bibinfo{year}{1970}).
\newblock


\bibitem[\protect\citeauthoryear{John and Srivastava}{John and
  Srivastava}{1999}]%
        {2-2}
\bibfield{author}{\bibinfo{person}{Oliver~P. John} {and}
  \bibinfo{person}{Sanjay Srivastava}.} \bibinfo{year}{1999}\natexlab{}.
\newblock \showarticletitle{The Big-Five trait taxonomy: History, measurement,
  and theoretical perspectives}.
\newblock \bibinfo{journal}{{\em Handbook of Personality\/}}
  \bibinfo{volume}{2} (\bibinfo{year}{1999}), \bibinfo{pages}{102--138}.
\newblock
\showURL{%
\url{http://fetzer.org/sites/default/files/images/stories/pdf/selfmeasures/Personality-BigFiveInventory.pdf}}


\bibitem[\protect\citeauthoryear{Kazis, Miller, Lee, Ren, Clark, Rogers,
  Spiro~III, Selim, Linzer, Payne, Mansell, and Fincke}{Kazis
  et~al\mbox{.}}{2006}]%
        {2-9}
\bibfield{author}{\bibinfo{person}{L.E. Kazis}, \bibinfo{person}{Skinner~K.M
  Miller, D.R.}, \bibinfo{person}{A. Lee}, \bibinfo{person}{X.S. Ren},
  \bibinfo{person}{J.A. Clark}, \bibinfo{person}{W.H. Rogers},
  \bibinfo{person}{A. Spiro~III}, \bibinfo{person}{A. Selim},
  \bibinfo{person}{M Linzer}, \bibinfo{person}{S.M. Payne}, \bibinfo{person}{D.
  Mansell}, {and} \bibinfo{person}{B.G. Fincke}.}
  \bibinfo{year}{2006}\natexlab{}.
\newblock \showarticletitle{Applications of Methodologies of the Veterans
  Health Study in the VA Health Care System: Conclusions and Summary}.
\newblock \bibinfo{journal}{{\em The Journal of Ambulatory Care Management\/}}
  \bibinfo{volume}{29}, \bibinfo{number}{2} (\bibinfo{year}{2006}),
  \bibinfo{pages}{182--188}.
\newblock
\showURL{%
\url{https://www.aaos.org/uploadedFiles/PreProduction/Quality/Measures/Veterans%20RAND%2012%20(VR-12).pdf}}


\bibitem[\protect\citeauthoryear{Kroenke, Spitzer, and Williams}{Kroenke
  et~al\mbox{.}}{2001}]%
        {2-6}
\bibfield{author}{\bibinfo{person}{Kurt Kroenke}, \bibinfo{person}{Robert~L.
  Spitzer}, {and} \bibinfo{person}{Janet~B.W. Williams}.}
  \bibinfo{year}{2001}\natexlab{}.
\newblock \showarticletitle{The PHQ-9: Validity of a Brief Depression Severity
  Measure}.
\newblock \bibinfo{journal}{{\em Journal of General Internal Medicine\/}}
  \bibinfo{volume}{16}, \bibinfo{number}{9} (\bibinfo{year}{2001}),
  \bibinfo{pages}{606--613}.
\newblock


\bibitem[\protect\citeauthoryear{Rui, Chen, Chen, Li, Harari, Tignor, Zhou,
  Ben-Zeev, and Campbell}{Rui et~al\mbox{.}}{2014}]%
        {2-1}
\bibfield{author}{\bibinfo{person}{Wang Rui}, \bibinfo{person}{Fanglin Chen},
  \bibinfo{person}{Zhenyu Chen}, \bibinfo{person}{Tianxing Li},
  \bibinfo{person}{Gabriella Harari}, \bibinfo{person}{Stefanie Tignor},
  \bibinfo{person}{Xia Zhou}, \bibinfo{person}{Dror Ben-Zeev}, {and}
  \bibinfo{person}{Andrew~T. Campbell}.} \bibinfo{year}{2014}\natexlab{}.
\newblock \showarticletitle{StudentLife: Assessing Mental Health, Academic
  Performance and Behavoiral Trends of College Students using Smartphones}.
\newblock \bibinfo{journal}{{\em In Proceedings of the ACM Conference on
  Ubiquitous Computing\/}} (\bibinfo{year}{2014}).
\newblock
\showURL{%
\url{http://studentlife.cs.dartmouth.edu/}}


\bibitem[\protect\citeauthoryear{Russell, Peplau, and Ferguson}{Russell
  et~al\mbox{.}}{1978}]%
        {2-4}
\bibfield{author}{\bibinfo{person}{Dan Russell}, \bibinfo{person}{Letitia~Anne
  Peplau}, {and} \bibinfo{person}{Mary~Lund Ferguson}.}
  \bibinfo{year}{1978}\natexlab{}.
\newblock \showarticletitle{Developing A Measure of Loneliness}.
\newblock \bibinfo{journal}{{\em Journal of Personality Assessment\/}}
  \bibinfo{volume}{42} (\bibinfo{year}{1978}), \bibinfo{pages}{290--294}.
\newblock
\showURL{%
\url{http://fetzer.org/sites/default/files/images/stories/pdf/selfmeasures/Self_Measures_for_Loneliness_and_Interpersonal_Problems_UCLA_LONELINESS.pdf}}


\bibitem[\protect\citeauthoryear{Sano, Phillips, Yu, McHill, Taylor, Jaques,
  Czeisler, Klerman, and Picard}{Sano et~al\mbox{.}}{2015}]%
        {3-2}
\bibfield{author}{\bibinfo{person}{Akane Sano}, \bibinfo{person}{Andrew~J.
  Phillips}, \bibinfo{person}{Amy~Z. Yu}, \bibinfo{person}{Andrew~W. McHill},
  \bibinfo{person}{Sara Taylor}, \bibinfo{person}{Natasha Jaques},
  \bibinfo{person}{Charles~A. Czeisler}, \bibinfo{person}{Elizabeth~B.
  Klerman}, {and} \bibinfo{person}{Rosalind~W. Picard}.}
  \bibinfo{year}{2015}\natexlab{}.
\newblock \showarticletitle{Recognizing Academic Performance, Sleep Quality,
  Stress Level, and Mental Health using Personality Traits, Wearable Sensors
  and Mobile Phones}.
\newblock \bibinfo{journal}{{\em 2015 International Conference on Wearable
  Implantable Body Senor Networks\/}} (\bibinfo{date}{June}
  \bibinfo{year}{2015}), \bibinfo{pages}{1--13}.
\newblock
\showDOI{%
\url{https://doi.org/10.1109/BSN.2015.7299420}}


\bibitem[\protect\citeauthoryear{Sapienza, Bessi, and Ferrara}{Sapienza
  et~al\mbox{.}}{2017}]%
        {sapienza2017non}
\bibfield{author}{\bibinfo{person}{Anna Sapienza}, \bibinfo{person}{Alessandro
  Bessi}, {and} \bibinfo{person}{Emilio Ferrara}.}
  \bibinfo{year}{2017}\natexlab{}.
\newblock \showarticletitle{Non-negative Tensor Factorization for Human
  Behavioral Pattern Mining in Online Games}.
\newblock \bibinfo{journal}{{\em arXiv preprint arXiv:1702.05695\/}}
  (\bibinfo{year}{2017}).
\newblock


\bibitem[\protect\citeauthoryear{Wang, Chen, Chen, Li, Harari, Tignor, Zhou,
  Ben-Zeev, and Campbell}{Wang et~al\mbox{.}}{2014}]%
        {3-1}
\bibfield{author}{\bibinfo{person}{Rui Wang}, \bibinfo{person}{Fanglin Chen},
  \bibinfo{person}{Zhenyu Chen}, \bibinfo{person}{Tianxing Li},
  \bibinfo{person}{Gabriella Harari}, \bibinfo{person}{Stefanie Tignor},
  \bibinfo{person}{Xia Zhou}, \bibinfo{person}{Dror Ben-Zeev}, {and}
  \bibinfo{person}{Andrew~T. Campbell}.} \bibinfo{year}{2014}\natexlab{}.
\newblock \showarticletitle{StudentLife: Assessing Mental Health, Academic
  Performance and Behavioral Trends of College Students using Smartphones}.
\newblock \bibinfo{journal}{{\em UbiComp '14 Proceedings of the 2014 ACM
  International Joint Conference on Pervasive and Ubiquitous Computing\/}}
  (\bibinfo{date}{September} \bibinfo{year}{2014}), \bibinfo{pages}{3--14}.
\newblock
\showURL{%
\url{10.1145/2632048.2632054}}


\bibitem[\protect\citeauthoryear{Wang, Harari, Hao, Zhou, and Campbell}{Wang
  et~al\mbox{.}}{2015}]%
        {wang2015smartgpa}
\bibfield{author}{\bibinfo{person}{Rui Wang}, \bibinfo{person}{Gabriella
  Harari}, \bibinfo{person}{Peilin Hao}, \bibinfo{person}{Xia Zhou}, {and}
  \bibinfo{person}{Andrew~T Campbell}.} \bibinfo{year}{2015}\natexlab{}.
\newblock \showarticletitle{SmartGPA: how smartphones can assess and predict
  academic performance of college students}. In \bibinfo{booktitle}{{\em
  Proceedings of the 2015 ACM international joint conference on pervasive and
  ubiquitous computing}}. ACM, \bibinfo{pages}{295--306}.
\newblock


\bibitem[\protect\citeauthoryear{Watson, Clark, and Tellegen}{Watson
  et~al\mbox{.}}{1988}]%
        {2-8}
\bibfield{author}{\bibinfo{person}{D. Watson}, \bibinfo{person}{L.A. Clark},
  {and} \bibinfo{person}{A. Tellegen}.} \bibinfo{year}{1988}\natexlab{}.
\newblock \showarticletitle{Development and Validation of Brief Measures of
  Positive and Negative Affect: The PANAS Scales}.
\newblock \bibinfo{journal}{{\em Journal of Personality and Social
  Psychology\/}} \bibinfo{volume}{54}, \bibinfo{number}{6}
  (\bibinfo{year}{1988}), \bibinfo{pages}{1063}.
\newblock
\showURL{%
\url{https://booksite.elsevier.com/9780123745170/Chapter%203/Chapter_3_Worksheet_3.1.pdf}}


\end{thebibliography}

\end{document}